# Designing Reliable LLM-Assisted Rubric Scoring for Constructed Responses: Evidence from Physics Exams

Xiuxiu Tang[1,2], G. Alex Ambrose[2], Ying Cheng[1]

[1] Department of Psychology, University of Notre Dame, Notre Dame, IN, 46556, USA
[2] Notre Dame Learning, University of Notre Dame, Notre Dame, IN, 46556, USA

Xiuxiu Tang: xtang8@nd.edu (corresponding author);
G. Alex Ambrose (gambrose@nd.edu); Ying Cheng (ycheng4@nd.edu)

## Abstract

Student responses in STEM assessments are often handwritten and combine symbolic expressions, calculations, and diagrams, creating substantial variation in format and interpretation. Despite their importance for evaluating students' reasoning, such responses are time-consuming to score and prone to rater inconsistency, particularly when partial credit is required. Recent advances in large language models (LLMs) have increased attention to AI-assisted scoring, yet evidence remains limited regarding how rubric design and LLM configurations influence reliability across performance levels. This study examined the reliability of AI-assisted scoring of undergraduate physics constructed responses using GPT-4o. Twenty authentic handwritten exam responses were scored across two rounds by four instructors and by the AI model using skill-based rubrics with differing levels of analytic granularity. Prompting format and temperature settings were systematically varied. Overall, human–AI agreement on total scores was comparable to human inter-rater reliability and was highest for high- and low-performing responses, but declined for mid-level responses involving partial or ambiguous reasoning. Criterion-level analyses showed stronger alignment for clearly defined conceptual skills than for extended procedural judgments. A more fine-grained, checklist-based rubric improved consistency relative to holistic scoring. These findings indicate that reliable AI-assisted scoring depends primarily on clear, well-structured rubrics, while prompting format plays a secondary role and temperature has relatively limited impact. More broadly, the study provides transferable design recommendations for implementing reliable LLM-assisted scoring in STEM contexts through skill-based rubrics and controlled LLM settings.

**Keywords:** AI-assisted scoring, inter-rater reliability, constructed responses, STEM assessment

# 1. Introduction

Constructed-response tasks play a central role in STEM education by eliciting students' conceptual reasoning and problem-solving processes (Neumann et al., 2013; Pellegrino, 2012). Such tasks provide rich evidence of student thinking, yet are time-consuming to score and susceptible to rater variability, particularly in high-enrollment gateway courses. Differences in rater severity, interpretation of ambiguous reasoning, and inconsistent application of scoring criteria can reduce score dependability, which in turn undermines the validity of instructional and placement decisions (Brookhart, 2013; Williamson et al., 2006).

Recent advances in generative artificial intelligence (AI), particularly large language models (LLMs), have renewed interest in automated approaches to scoring constructed responses. Compared with earlier feature-based scoring systems, LLMs can process extended responses, identify patterns of reasoning, and generate natural-language evaluations aligned with scoring criteria (Floridi & Chiriatti, 2020; Kasneci et al., 2023). Empirical studies suggest that, when prompts and rubrics are clearly specified, LLMs can approximate human scoring on essay-based tasks (e.g., Tate et al., 2024; Quah et al., 2024; Liu et al., 2024; Yavuz et al., 2025). This emerging evidence suggests that generative AI may contribute to more efficient assessment and feedback processes.

Extending these results to STEM constructed responses, however, remains challenging. Student work in physics and related disciplines often combines symbolic expressions, intermediate calculations, diagrams, and handwritten annotations, with substantial variation in notation, legibility, and spatial organization. Such responses may include informal symbols, crossed-out work, or incomplete solution paths, all of which complicate consistent interpretation. Prior research shows that this multimodal and variably formatted work poses persistent challenges for AI-based scoring systems, which may struggle to apply scoring criteria reliably across representations (Caraeni et al., 2024; Kortemeyer et al., 2024; Mok et al., 2025).

Beyond these practical challenges, several empirical gaps remain in the literature. First, much existing research on AI-assisted scoring emphasizes holistic judgments rather than analytic, multi-criterion scoring, which is essential for diagnosing specific strengths and misconceptions in STEM learning (Holmes & Wieman, 2016; Reynders et al., 2020). When rubrics rely primarily on holistic descriptors or loosely specified criteria, both human raters and AI must infer evaluative standards from incomplete guidance. This ambiguity is most consequential for partially correct responses, where multiple reasoning components must be weighed and integrated.

Second, prior studies rarely examine how agreement between AI and human raters varies across performance levels. Most evaluations rely on overall correlations or average agreement indices (e.g., Bui and Barrot, 2025; Kortemeyer et al., 2024). Such aggregate analyses may obscure systematic differences in scoring accuracy for high-, mid-, and low-performing students. In practice, partial-credit responses from mid-performing students frequently require the most interpretive judgment and are most susceptible to rater disagreement. If AI performs differently across performance levels, failure to examine these patterns may mask sources of variation and threaten the fairness and validity of resulting scores.

Third, limited attention has been given to how the configurations of LLMs influence performance, including prompting strategies and temperature settings, both influencing the model outputs. Without systematic evidence on how these parameters interact with rubric structure and performance level, it remains difficult to develop principled guidelines for classroom deployment.

Overall, these gaps indicate a lack of empirical evidence on how rubric design and LLM configurations shape the performance of AI-assisted scoring across performance levels in authentic STEM assessment contexts. Accordingly, this study examines how different levels of rubric granularity influence scoring consistency. Rubric development was informed by principles from cognitive diagnostic modelling (CDM), which emphasize decomposing problems into underlying skills required for successful problem solving (Rupp et al., 2010). Building on this foundation, we implemented two successive CDM-informed rubrics that differed in analytic granularity. In Round 1, skills were scored holistically across the full response. In Round 2, the rubric was refined into a more fine-grained, checklist-style format with explicit, part-specific criteria. This iterative refinement reflects a design-oriented effort to reduce interpretive variability and improve the reliability of partial-credit evaluation. In addition, we examined the reliability of AI-assisted scoring under systematically varied prompting conditions (0-shot vs. 1-shot) and temperature settings (0.3, 0.5, 0.7, and 1.0). Across two rounds, four experienced instructors independently evaluated student work using the CDM-informed rubrics, and their judgments were compared with AI-generated scores. Using intraclass correlation coefficients (ICCs), we assessed agreement at both the total-score and criterion levels and examined how reliability varied across performance groups.

As such, this study provides empirical evidence on how rubric design and LLM configurations influence the reliability of AI-assisted scoring across performance levels. The findings offer practical guidance for instructors and institutions seeking to integrate generative AI into assessment workflows while supporting principled evaluation and meaningful formative feedback in STEM education.

## 2. Literature Review

## 2.1 Foundations of Automated Scoring in Education

Automated scoring has long been explored as a way to reduce the time, cost, and inconsistency associated with human evaluation of open-ended student work. Human raters often vary in severity, interpretation, and susceptibility to fatigue, which can introduce unwanted score variability and compromise the dependability of results in high-enrollment or high-stakes settings (Brookhart, 2013; Williamson et.al., 2006). Early automated essay scoring (AES) systems such as *e-rater* and *IntelliMetric* attempted to address these challenges by predicting human scores using hand-engineered linguistic, syntactic, and mechanical features (Attali & Burstein, 2006; Rudner & Liang, 2002; Shermis & Burstein, 2013). These systems achieved levels of agreement comparable to trained human raters and were adopted in several large-scale assessment programs, including the Test of English as a Foreign Language (TOEFL), the Graduate Record Examinations (GRE), and the Graduate Management Admission Test (GMAT) (Powers et al., 2002; Wang & von Davier, 2014; Rudner & Liang, 2002; Shermis & Burstein, 2013).

Despite their operational success, early AES systems were criticized for relying on surface-level textual features that correlated with writing quality without necessarily capturing deeper conceptual understanding. Empirical studies demonstrated that some systems rewarded response length or advanced vocabulary independently of substantive content (Perelman, 2014; Deane, 2013). From an educational measurement perspective, these behaviors raised fundamental validity concerns related to construct representation, interpretability, and alignment with instructional goals (Messick, 1995; Kane, 2013). Subsequent neural network–based scoring models improved predictive performance by learning latent text representations directly from data (Taghipour & Ng, 2016; Alikaniotis et al., 2016), but concerns about transparency and construct representation persisted. These developments highlight a persistent tension in automated scoring between predictive accuracy and measurement validity. Advances in generative AI represent a new methodological approach that differs fundamentally from earlier feature-based and predictive scoring systems. The extent to which these newer models address longstanding measurement challenges while introducing new sources of variability is an important question for contemporary assessment research.

**2.2 Generative AI as a Scoring Tool: Capabilities and Challenges**

Generative AI has introduced new possibilities for automated scoring of open-ended student work (Floridi & Chiriatti, 2020; Achiam et al., 2023). Unlike earlier feature-based AES systems, LLMs can process extended, context-rich responses and generate text that is conditioned on the full content of the submission. With appropriate prompting, they can produce rubric-aligned scores and brief explanations that correspond to the content presented in the response. This prompting-based approach allows scoring to be guided through natural language instructions rather than fixed feature sets, which increases flexibility and supports adaptation to a broad range of tasks and rubrics.

Empirical evidence shows that LLMs can achieve moderate to strong alignment with trained human raters when the scoring task is clearly specified and the prompt includes structured rubric information. For example, Tate et al. (2024) found that GPT-4 generated holistic essay scores that closely matched those of trained human raters across a large dataset of student essays. Similarly, Quah et al. (2024) found that GPT-4's automated essay scores showed moderate to high agreement with instructor evaluations in a dental education context. These results suggest that LLMs may be suitable for initial scoring or low-stakes feedback in educational settings.

However, several limitations remain. LLM outputs are sensitive to prompt design, rubric phrasing, and contextual framing, with small changes in instructional language leading to different scoring outcomes (Mansour et al., 2024; Tian et al., 2024; Li & Liu, 2024). Changes to models introduce additional instability; the same response may receive different scores over time as model updates alter underlying behavior. For example, the study by Bui and Barrot (2025) discovered that ChatGPT failed to provide similar scores for the same essays when graded at two different points in time.

In addition, research from the LLM-as-judge literature documents systematic scoring tendencies, including verbosity bias and positional effects that favor certain responses independent of substantive quality (Hu et al., 2024; Shi et al., 2024; Wu & Aji, 2025). Although these studies are not situated directly in educational assessment, they raise concerns about latent heuristics that may influence AI-assisted scoring. Furthermore, although LLMs generate coherent explanations for their scores, recent work shows that these rationales may not faithfully represent the processes underlying the assigned scores and can obscure reliance on heuristic shortcuts (Turpin et al., 2023). These challenges suggest the need for empirical validation of AI-assisted scoring across varied tasks and prompting conditions.

## 2.3 AI-Assisted Scoring in STEM Assessment

Interest in AI-assisted scoring has grown within STEM education, where constructed responses frequently integrate conceptual explanation, symbolic manipulation, graphical representation, and multi-step reasoning. Such responses differ markedly from linear, text-based essays and pose distinctive challenges for automated interpretation. A recent review found that AI-based grading systems can reduce instructor workload and improve the timeliness of feedback, yet their accuracy varies substantially across item types, response structures, and instructional contexts (Tan et al., 2025).

Prior research demonstrates that high human–AI scoring agreement in STEM is achievable for constructed-response items under appropriate conditions, particularly when responses are clearly structured and scoring criteria are explicit. For instance, Morris et al. (2024) reported that an LLM-based scoring model achieved human-like agreement with raters for nine of ten NAEP mathematics items, with lower agreement observed for items involving ambiguous or context-dependent response formats that were inconsistently interpreted by human raters. Similarly, Kortemeyer et al. (2024) found that AI-assisted grading of handwritten thermodynamics exams achieved reasonable alignment with instructor scores when textual representations were reliable, but performance degraded when item responses involved diagrams, informal notation, unconventional layouts, or partially correct solution paths. In these cases, limitations in optical character recognition and response representation propagated downstream errors into the grading process.

Rubric structure also plays a critical role in STEM assessment, where analytic rubrics are commonly used to differentiate conceptual understanding, procedural execution, representational accuracy, and explanation quality. Studies have shown that LLM-based scoring aligns more closely with human judgment when explicit mark schemes or structured rubrics are provided, whereas scoring without such guidance can be inconsistent or overly lenient (Mok et al., 2025). Beyond general analytic scoring, some assessment approaches emphasize decomposing performance into interpretable skill or knowledge components. CDM provides a theoretical framework for conceptualizing such skill-based representations by linking observable response features to underlying attributes or competencies (Rupp et al., 2010). Although CDM was originally developed as a psychometric modeling approach rather than a rubric-design methodology, its emphasis on explicit skill-level representations offers a useful conceptual basis for designing analytic rubrics that support structured and interpretable scoring. In this study, CDM concepts are used to develop analytic rubrics that enable skill-level scoring of physics constructed responses.

**Research Questions**

To guide this investigation, we address the following research questions:

1) **RQ1.** To what extent do AI-generated scores agree with scores assigned by experienced instructors using skill-based rubrics, at both the total-score and criterion levels?
2) **RQ2.** Does human–AI scoring agreement vary across student performance levels under different rubric structures?
3) **RQ3.** How do rubrics, prompting format and temperature settings affect human–AI scoring agreement in undergraduate physics assessment?

### 3. Methods

**3.1 Participants and Context**

The study was conducted in a large-enrollment, required physics course for first-year engineering students at a U.S. university. Twenty authentic handwritten student responses to a three-part computational exam problem were collected (see Figure 1 for the exam problem and Figure 2 for an example student response). The sample was purposefully selected to include high-, mid-, and low-performing responses in order to examine scoring reliability across performance levels.

The problem consisted of three related sub-questions that required students to apply kinematic equations and vector decomposition across successive stages of a projectile motion scenario. Responses were produced under standard exam conditions and reflected authentic problem-solving approaches. All sample responses were scanned and stored as digital images for subsequent AI-assisted scoring.

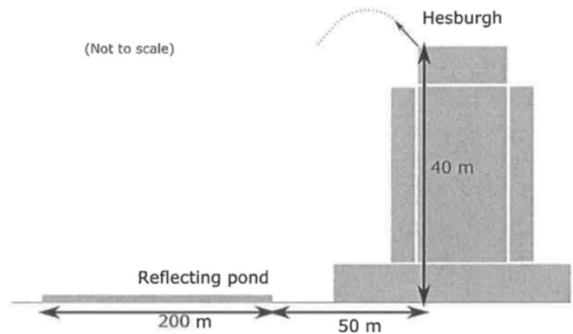

Figure 1. The Selected Physics Exam Problem

Figure 2. An Example Student Response

To balance feasibility with analytic depth, the dataset was divided into two sets of ten unique responses that were analyzed in two sequential scoring rounds. The two rounds differed in rubric structure and AI scoring conditions, allowing systematic examination of how scoring reliability depends on rubric design and evaluation strategy. Four experienced physics instructors, all of whom had previously taught and graded in the course, collaboratively developed both rubrics and served as independent human raters in both rounds.

**3.2 Rubric Development**

Rubric development was informed by principles commonly associated with CDM, in which complex problem-solving performance is conceptualized as a combination of underlying skills. In CDM-based approaches, item responses are interpreted with reference to a predefined skill structure, often represented through a Q-matrix that specifies which skills are required for successful performance. Following this framework, the four instructors deconstructed the exam problem into a set of core skills required for successful solution and organized these skills into observable scoring criteria. Two rubric versions were developed and applied sequentially across the two rounds to support different levels of scoring granularity. Both rubrics were skill-based; their primary distinction lay in how skills were operationalized during scoring.

*Rubric Version 1*

Rubric version 1 (V1), used in Round 1, was a skill-based rubric scored holistically across problem parts (see Table A1 in the Appendix). The rubric comprised six skill criteria spanning three domains—Physics Knowledge, Math Skills, and Problem Solving. Each skill was scored on a four-point scale (Mastery [3], Growing [2], Attempted [1], and No Attempt [0]) with qualitative descriptors defining expectations at each level.

The six skills assessed in V1 included recognition of constant acceleration, selection of appropriate projectile equations, conversion between vector magnitude–angle and component forms, extraction of relevant problem information, identification of applicable principles or equations, and execution of mathematical procedures. For each skill, raters reviewed the student's complete solution across all three parts of the problem and assigned a single score reflecting overall performance on that skill. Skill scores were therefore informed by evidence distributed across multiple parts of the response but were not tied to part-specific scoring rules or discrete reasoning steps. Partial credit judgments relied on rater interpretation guided by qualitative descriptors.

### *Rubric Version 2*

Rubric version 2 (V2), used in Round 2, was a skill-based checklist rubric designed to reduce ambiguity in partial-credit decisions and support more fine-grained reliability analysis (see Table A2 in the Appendix). V2 retained the same three overarching domains as V1 but redefined the skill structure within those domains to better align with observable reasoning steps and part-specific evidence. While V2 also comprised six skill criteria, these skills were not intended to be a one-to-one mapping from V1; instead, they reflected a refined decomposition of problem-solving processes suited to checklist-based scoring.

In V2, each skill was operationalized using explicit, part-specific indicators, and points were awarded for discrete reasoning steps, intermediate results, and correct final answers within each sub-question. For example, Physics Knowledge criteria differentiated correct application of vertical motion equations in part (a), coordinated use of horizontal and vertical components in part (b), and interpretation of projectile outcomes in part (c). Problem-solving skills were subdivided into extracting relevant information, identifying governing principles, converting among representational forms, and executing algebraic or numerical procedures, with each component evaluated using clearly specified checklist items. By anchoring scores to explicit features of student work and allowing multiple opportunities to demonstrate related skills within a single response, V2 reduced subjectivity in partial-credit scoring and enabled more precise analysis of scoring reliability across skills.

### 3.3 Scoring Process

*Human Scoring Procedure*

All student responses were independently scored by the four physics instructors. In each round, the instructors first participated in a calibration session in which they reviewed the rubric and practiced scoring sample responses not included in the dataset. This calibration process supported a shared interpretation of scoring criteria and performance descriptors. After calibration, each instructor independently scored all responses in their assigned round using the corresponding rubric. No discussion occurred during scoring, and raters did not have access to AI-generated scores. Each response therefore received four independent human ratings in both rounds.

*AI Scoring Procedure*

Automated scoring was conducted using GPT-4o, a multimodal LLM developed by OpenAI. GPT-4o can process handwritten images directly, allowing student responses to be scored in their original form without transcription or optical character recognition. In both rounds, the model was instructed to assign scores using the same rubric applied by the human raters. In Round 1, AI scoring was conducted once per response using a single prompt consisting of the scoring rubric and instructions, without any scored examples (i.e., 0-shot prompting). The temperature parameter was set to 1.0, which corresponds to the model's default setting and allows relatively unconstrained response generation. This round was intended to evaluate baseline AI scoring performance under typical default conditions.

In Round 2, AI scoring was expanded to systematically vary two parameters: prompt format (0-shot vs. 1-shot) and temperature. Under the 0-shot condition, the prompt consisted solely of the scoring rubric and explicit instructions to apply the rubric to the student response. Under the 1-shot condition, the same rubric and instructions were supplemented with one example student response that received full credit, along with the corresponding scores assigned by four human raters using the rubric, to illustrate how the rubric should be applied. Temperature was set to 0.3, 0.5, 0.7, and 1.0 to control the degree of randomness in the model's output. Lower temperature values produce more deterministic and consistent responses, whereas higher values introduce greater variability, making the model's scoring less constrained and more sensitive to alternative interpretations. Each combination of prompt format and temperature was repeated four times per response to assess the stability of AI-generated scores under identical settings.

**3.4 Reliability and Data Analysis**

Reliability was evaluated using intraclass correlation coefficients (ICCs) estimated under a two-way random-effects model with absolute agreement (Shrout & Fleiss,1979; McGraw & Wong, 1996). Five raters were included in the analysis: four human raters and GPT-4o, which was treated as an additional scoring agent applying the same rubric. In this model, responses and raters are treated as random effects, meaning that the human raters represent a sample from a larger population of qualified raters. The resulting ICC reflects the reliability of the average score across all raters and the extent to which these scores would generalize beyond the specific human raters included in the study. ICCs quantify the extent to which different raters assign the same scores to the same responses and are widely used to assess inter-rater reliability in educational assessment. Consistent with common conventions, ICC values below 0.50 were interpreted as poor, values from 0.50 to 0.75 as moderate, values from 0.75 to 0.90 as good, and values above 0.90 as excellent (Koo & Li, 2016).

Analyses were conducted at multiple levels. First, overall ICCs were computed to evaluate agreement on total scores. Second, ICCs were estimated separately for performance-level subgroups to compare agreement for extreme performers, defined as responses with very high or very low total scores, and mid-level performers. Third, criterion-level ICCs were calculated to assess agreement within individual rubric components. For each level of analysis, two sets of ICCs were reported: human-only ICCs, based on scores from the four human raters, and human–AI ICCs, in which AI-generated scores were included as an additional rater. Human-only ICCs established baseline inter-rater reliability, while human–AI ICCs evaluated the extent to which AI scoring aligned with human judgment. In addition to reliability analyses, descriptive statistics, including means and standard deviations, were computed to characterize score distributions and variability across raters, prompting formats, and temperature settings.

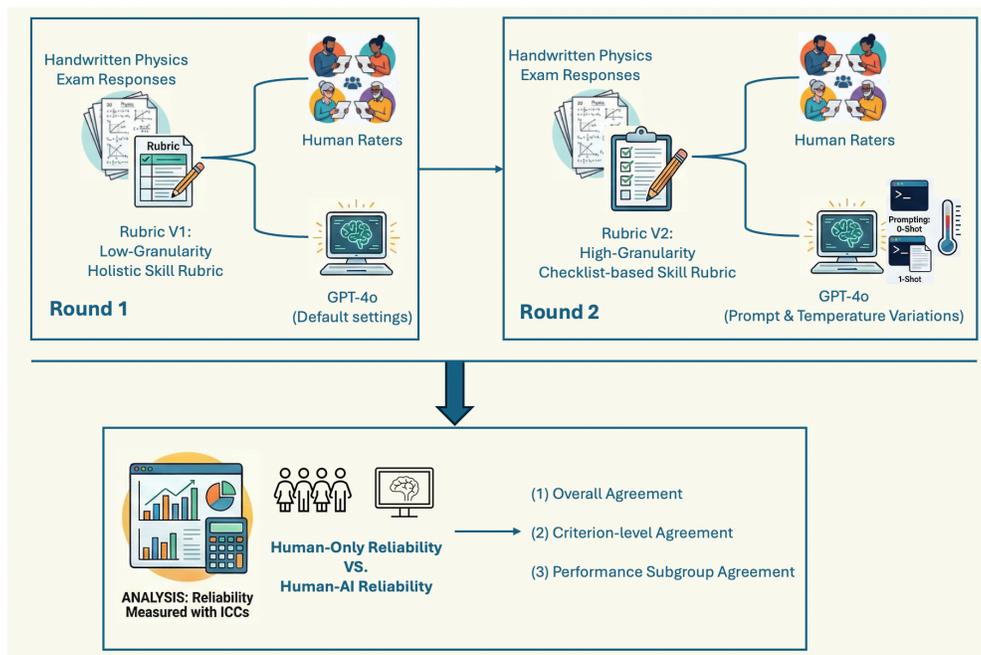

*Figure 3. Study Workflow Overview*

## 3.5 Workflow Summary

The study followed a sequential design. In Round 1, responses were scored by human raters and GPT-4o using rubric V1 under default conditions. Findings from this round informed the development of rubric V2, which emphasized checklist-based scoring. In Round 2, responses were again scored by human raters and GPT-4o, with systematic variation in prompt format and temperature. Reliability analyses were conducted at overall, criterion, and performance levels. Figure 3 provides an overview of the study workflow.

## 4. Results

### 4.1 Round 1: Scoring with Low-Granularity Holistic Skill Rubric

#### 4.1.1 Descriptive Comparisons

Table 1 and Figure 4 summarize scoring patterns for Round 1. Across the 10 responses, variability among the four human raters was modest, with within-student standard deviations (SDs) ranging from 0 to 2.36 points, indicating generally consistent scoring. GPT-4o's scores were close to the mean of human ratings for most students. For example, Student 1 received a perfect score of 18 from all four raters and from GPT-4o, while Student 9 was consistently scored near the lower end of the scale (human mean = 5.00, SD = 1.63; GPT-4o = 5).

Table 1. Scoring Results in Round 1

|  | Rater 1 | Rater 2 | Rater 3 | Rater 4 | Mean Rater Scores | SD of Rater Scores | GPT-4o |
|---|---|---|---|---|---|---|---|
| Student 1 | 18 | 18 | 18 | 18 | 18.00 | 0.00 | 18 |
| Student 2 | 7 | 5 | 6 | 1 | 4.75 | 2.63 | 6 |
| Student 3 | 17 | 17 | 17 | 14 | 16.25 | 1.50 | 18 |
| Student 4 | 12 | 14 | 11 | 9 | 11.50 | 2.08 | 13 |
| Student 5 | 17 | 18 | 16 | 18 | 17.25 | 0.96 | 18 |
| Student 6 | 16 | 13 | 13 | 12 | 13.50 | 1.73 | 11 |
| Student 7 | 15 | 17 | 15 | 20 | 16.75 | 2.36 | 18 |
| Student 8 | 17 | 17 | 15 | 18 | 16.75 | 1.26 | 18 |
| Student 9 | 5 | 5 | 7 | 3 | 5.00 | 1.63 | 5 |
| Student 10 | 16 | 16 | 14 | 11 | 14.25 | 2.36 | 18 |

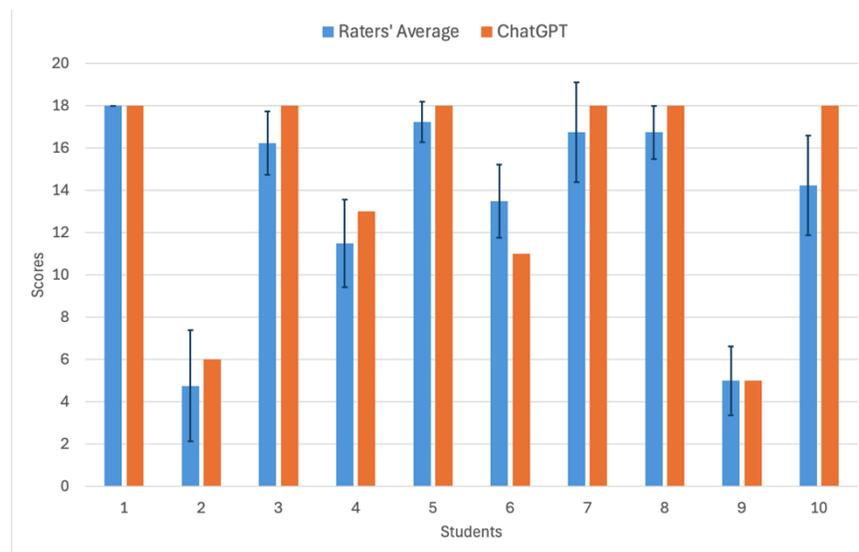

Figure 4. Scoring Results in Round 1

Differences between human and AI scores were more apparent for responses in the middle of the score range. Fo instance, for Student 4, human ratings ranged from 9 to 14 (mean = 11.50, SD = 2.08), while GPT-4o assigned a score of 13. Similarly, student 6 received human scores ranging from 12 to 16 (mean = 13.50, SD = 1.73), whereas GPT-4o assigned a score of 11, slightly below the human range. In most cases, GPT-4o's scores fell within the range of human ratings or remained close to the boundary of the human scoring range.

**4.1.2 Reliability**

*Overall Reliability*

As shown in Table 2, inter-rater reliability among the four human raters in Round 1 was good, with an ICC of 0.88, indicating consistent application of the holistic rubric. When GPT-4o's scores were included as an additional rater, the overall ICC increased slightly to 0.89, and the value comparable to the human-only ICC, indicating that GPT-4o's scores were broadly consistent with human ratings at the aggregate level.

Table 2. Overall and Performance Level ICCs in Round 1

|  | Overall (N = 10) | Extreme Performers (N = 7) | Mid-level Performers (N = 3) |
|---|---|---|---|
| Human Only | 0.88 | 0.92 | 0.28 |
| Human & AI (GPT-4o) | 0.89 | 0.94 | 0.30 |

*Reliability by Performance Level*

Table 2 also reports inter-rater reliability separately for extreme and mid-level performers. For extreme performers, inter-rater reliability was excellent among human raters (ICC = 0.92). Including GPT-4o as an additional rater yielded a similarly high ICC of 0.94, indicating strong agreement across raters for responses that were clearly correct or clearly incorrect.

In contrast, reliability was substantially lower for mid-level performers. Human-only agreement for this group was poor, with an ICC of 0.28. When GPT-4o's scores were included, the ICC increased slightly to 0.30 but remained low. These results indicate that responses reflecting partial understanding posed challenges for consistent scoring, with GPT-4o largely reproducing the variability already present in human judgments.

*Reliability by Scoring Criterion*

Criterion-level ICCs for Round 1 are reported in Table 3. Agreement varied systematically across scoring criteria. For criterion A1, which assessed recognition of kinematic concepts, inter-rater reliability was excellent among human raters (ICC = 0.93) and remained high when GPT-4o was included (ICC = 0.90). Similarly, criterion A2, which focused on projectile trajectory, showed good reliability for both humans (ICC = 0.84) and the combined human–AI ratings (ICC = 0.83). Criterion A3, assessing vector conversion, also demonstrated high agreement, with ICCs of 0.89 for human raters and 0.82 when GPT-4o was included.

Lower levels of agreement were observed for criteria involving more interpretive judgment. For criterion A4, which evaluated extraction of relevant problem information, reliability was moderate among human raters (ICC = 0.63) and slightly higher when GPT-4o was included (ICC = 0.68). Agreement declined further for criteria A5 and A6. For criterion A5, which assessed identification of appropriate principles, the human-only ICC was 0.44 and increased to 0.50 with the inclusion of GPT-4o. For criterion A6, which focused on mathematical procedures, reliability was similarly low, with ICCs of 0.41 for human raters and 0.51 for the combined human–AI ratings.

Table 3. Criterion-level ICCs in Round 1

| Scoring Criteria | Humans only | Humans and GPT-4o |
|---|---|---|
| A1 (Recognizing kinematics) | 0.93 | 0.90 |
| A2 (Projectile trajectory) | 0.84 | 0.83 |
| A3 (Vector conversion) | 0.89 | 0.82 |
| A4 (Extracting problem information) | 0.63 | 0.68 |
| A5 (Identifying principles) | 0.44 | 0.50 |
| A6 (Mathematical procedures) | 0.41 | 0.51 |

Overall, higher agreement was observed for criteria associated with clearly defined conceptual knowledge and representational transformations, whereas lower agreement was observed for criteria associated with interpretive or procedural judgments. Across criteria, the inclusion of GPT-4o produced ICC values that were comparable to those obtained from human raters alone.

**4.2 Round 2: Scoring with High-Granularity Checklist-based Skill Rubric**

**4.2.1 Descriptive Comparisons**

In Round 2, student responses were evaluated using Rubric V2, a checklist-based skill rubric that supported fine-grained scoring across multiple sub-criteria. Each response was scored by four human raters and by GPT-4o under systematically varied prompting conditions (0-shot and 1-shot) and temperature settings (0.3, 0.5, 0.7, and 1.0). Tables 4 and 5 report the mean and SD of scores across conditions, with Figures 5 and 6 providing visual comparisons. As in Round 1, human agreement was highest for responses demonstrating clear mastery (e.g., Students 13–15), while variability increased for responses with greater interpretive ambiguity.

Under 0-shot prompting, GPT-4o's scores closely tracked human scoring patterns across most students. For high-performing responses, AI scores were consistently near the maximum and aligned closely with human means across temperature settings (e.g., Students 13–16). Greater divergence emerged for mid-level responses, where human variability was higher and GPT-4o's mean scores deviated more from human averages (e.g., Student 17, human mean = 14.50, SD = 2.38; AI means = 9.25–13.00).

Prompting format exerted a stronger influence on AI scoring than temperature. Under 1-shot prompting, GPT-4o continued to align with human ratings for high-performing responses but showed substantial divergence for mid- and lower-performing responses, with both inflated and deflated scores relative to human means. In contrast, differences across temperature settings were relatively small and did not follow a consistent monotonic pattern.

Overall, GPT-4o closely reproduced human scoring for responses with low interpretive ambiguity, particularly under 0-shot prompting, while greater divergence and variability were observed for mid-level responses under 1-shot prompting.

Table 4. Round 2 Scoring Results (0-shot)

| | Human Raters | | AI (GPT-4o), 0-shot | | | | | | | |
| | | | T = 0.3 | | T = 0.5 | | T = 0.7 | | T = 1.0 | |
| Student | Mean | SD | Mean | SD | Mean | SD | Mean | SD | Mean | SD |
|---|---|---|---|---|---|---|---|---|---|---|
| 11 | 16.25 | 0.96 | 17.25 | 0.50 | 16.75 | 1.71 | 17.50 | 1.73 | 14.75 | 0.96 |
| 12 | 17.75 | 1.26 | 20.00 | 0.00 | 20.00 | 0.00 | 20.00 | 0.00 | 19.75 | 0.50 |
| 13 | 20.00 | 0.00 | 20.00 | 0.00 | 19.00 | 2.00 | 20.00 | 0.00 | 19.00 | 0.82 |
| 14 | 20.00 | 0.00 | 20.00 | 0.00 | 20.00 | 0.00 | 20.00 | 0.00 | 20.00 | 0.00 |
| 15 | 20.00 | 0.00 | 20.00 | 0.00 | 20.00 | 0.00 | 19.75 | 0.50 | 20.00 | 0.00 |
| 16 | 18.25 | 0.96 | 19.75 | 0.50 | 19.75 | 0.50 | 19.00 | 1.15 | 18.50 | 1.00 |

| Student | | | | | | | | | | |
|---|---|---|---|---|---|---|---|---|---|---|
| 17 | 14.50 | 2.38 | 13.00 | 3.92 | 9.25 | 4.27 | 10.25 | 3.59 | 9.50 | 3.11 |
| 18 | 18.00 | 0.82 | 20.00 | 0.00 | 20.00 | 0.00 | 20.00 | 0.00 | 19.75 | 0.50 |
| 19 | 6.00 | 2.71 | 11.25 | 0.96 | 11.50 | 1.29 | 11.75 | 1.26 | 8.25 | 3.77 |
| 20 | 6.75 | 1.26 | 7.50 | 1.29 | 6.50 | 3.32 | 9.75 | 2.63 | 7.25 | 2.63 |

*Note.* T = temperature.

Table 5. Round 2 Scoring Results (1-shot)

| | Human Raters | | AI (GPT-4o), 1-shot | | | | | | | |
|---|---|---|---|---|---|---|---|---|---|---|
| | | | T = 0.3 | | T = 0.5 | | T = 0.7 | | T = 1.0 | |
| Student | Mean | SD | Mean | SD | Mean | SD | Mean | SD | Mean | SD |
| 11 | 16.25 | 0.96 | 17.00 | 0.00 | 17.75 | 2.63 | 14.00 | 6.06 | 11.50 | 1.29 |
| 12 | 17.75 | 1.26 | 20.00 | 0.00 | 20.00 | 0.00 | 19.25 | 0.96 | 18.00 | 1.83 |
| 13 | 20.00 | 0.00 | 20.00 | 0.00 | 20.00 | 0.00 | 20.00 | 0.00 | 20.00 | 0.00 |
| 14 | 20.00 | 0.00 | 20.00 | 0.00 | 20.00 | 0.00 | 20.00 | 0.00 | 20.00 | 0.00 |
| 15 | 20.00 | 0.00 | 20.00 | 0.00 | 20.00 | 0.00 | 20.00 | 0.00 | 20.00 | 0.00 |
| 16 | 18.25 | 0.96 | 20.00 | 0.00 | 19.75 | 0.50 | 20.00 | 0.00 | 19.75 | 0.50 |
| 17 | 14.50 | 2.38 | 3.00 | 1.63 | 3.25 | 2.63 | 3.00 | 2.83 | 6.00 | 1.63 |
| 18 | 18.00 | 0.82 | 10.50 | 6.45 | 8.75 | 2.22 | 10.50 | 4.20 | 12.50 | 6.14 |
| 19 | 6.00 | 2.71 | 5.50 | 0.58 | 3.75 | 1.71 | 3.25 | 1.50 | 6.25 | 2.63 |
| 20 | 6.75 | 1.26 | 13.25 | 6.24 | 12.50 | 6.24 | 15.50 | 7.14 | 15.75 | 5.85 |

*Note.* T = temperature.

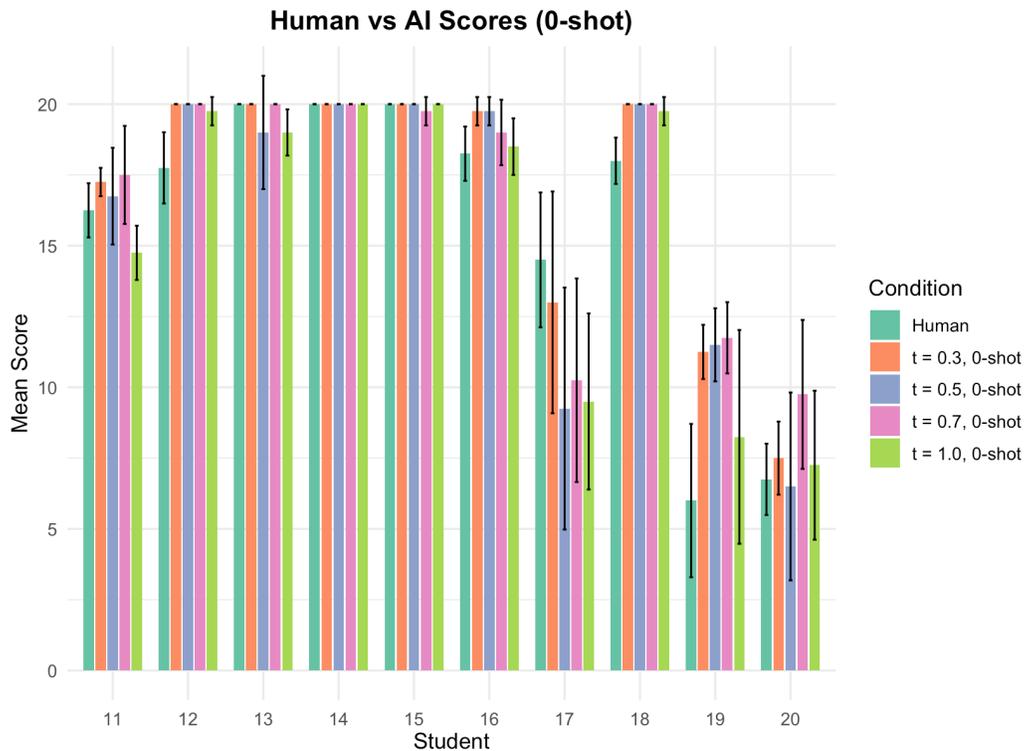

Figure 5. Human vs AI (GPT-4o) Scores (0-shot) in Round 2

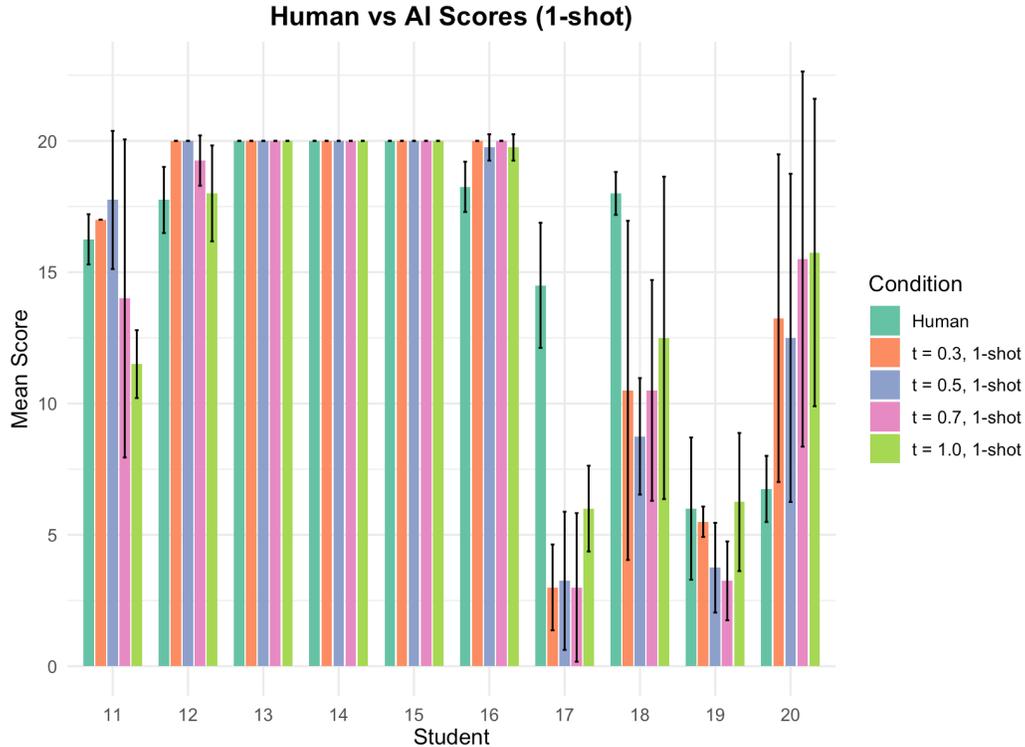

Figure 6. Human vs AI (GPT-4o) Scores (1-shot) in Round 2

**4.2.2 Overall Reliability**

As shown in Table 6, the overall inter-rater reliability among the four human raters in Round 2 was excellent, with an ICC of 0.94, indicating highly consistent application of the checklist rubric. When GPT-4o scores generated under 0-shot prompting were included as an additional rater, overall reliability remained high, with ICC values ranging from 0.90 to 0.92 across temperature settings. These values fall within the same reliability range as the human-only estimate, indicating close overall agreement between human raters and GPT-4o under 0-shot conditions.

In contrast, overall reliability estimates were lower and more variable when GPT-4o scores generated under 1-shot prompting were included. ICC values under 1-shot prompting ranged from 0.78 to 0.92 across temperature settings. While some conditions continued to yield high agreement, overall reliability was less stable under 1-shot prompting than under 0-shot prompting. Across both prompting formats, differences in ICC values across temperature settings were relatively small, suggesting that temperature had a limited effect on overall reliability compared with prompting format.

Table 6. Overall and Performance Level ICCs in Round 2

|  | Temperature | Overall | Extreme Performers | Mid-level Performers |
|---|---|---|---|---|
| Human only | - | 0.94 | 0.94 | 0.89 |
| Human & AI (GPT-4o), 0-shot | T = 0.3 | 0.92 | 0.90 | 0.90 |
|  | T = 0.5 | 0.90 | 0.89 | 0.80 |
|  | T = 0.7 | 0.90 | 0.89 | 0.77 |
|  | T = 1.0 | 0.92 | 0.93 | 0.80 |
| Human & AI (GPT-4o), 1-shot | T = 0.3 | 0.80 | 0.89 | 0.07 |
|  | T = 0.5 | 0.80 | 0.86 | 0.14 |
|  | T = 0.7 | 0.78 | 0.89 | -0.09 |
|  | T = 1.0 | 0.92 | 0.90 | 0.11 |

### 4.2.3 Reliability by Performance Level

Table 6 also reports inter-rater reliability separately for extreme and mid-level performers in Round 2. Among human raters, reliability was high for both performance groups, with ICCs of 0.94 for extreme performers and 0.89 for mid-level performers, indicating consistent application of the checklist rubric across response types.

When GPT-4o scores generated under 0-shot prompting were included, reliability for extreme performers remained high across temperature settings, with ICCs ranging from 0.89 to 0.93. For mid-level performers, ICCs under 0-shot prompting ranged from 0.77 to 0.90, indicating moderate to good agreement, with some variability across temperature settings. Overall, reliability for mid-level performers under 0-shot prompting remained substantially higher than that observed in Round 1.

A different pattern emerged under 1-shot prompting. For extreme performers, ICCs remained high across temperatures, ranging from 0.86 to 0.90. In contrast, reliability for mid-level performers dropped sharply under 1-shot prompting, with ICCs ranging from −0.09 to 0.14 across temperature settings. These values indicate little to no agreement between human raters and GPT-4o for mid-level responses under 1-shot conditions.

Across both prompting formats, differences in ICCs across temperature settings were relatively small for extreme performers, whereas substantial variability was observed for mid-level performers under 1-shot prompting. These results indicate that agreement between human raters and GPT-4o in Round 2 depended more strongly on prompting format and performance level than on temperature setting.

**4.2.4 Reliability by Scoring Criterion**

Criterion-level ICCs for Round 2 are reported in Table 7 and illustrated in Figures 7 and 8. Agreement varied across scoring criteria and prompting conditions. For criterion A1 (physics knowledge), reliability was excellent among human raters (ICC = 0.93) and remained high when GPT-4o scores generated under 0-shot prompting were included, with ICCs ranging from 0.91 to 0.92 across temperature settings. Under 1-shot prompting, reliability for A1 was lower but remained in the good range, with ICCs between 0.80 and 0.83.

Criterion A2 (math skill) also showed good agreement. Human-only reliability was high (ICC = 0.89), and ICCs under 0-shot prompting ranged from 0.82 to 0.85. Under 1-shot prompting, reliability for A2 declined modestly, with ICCs ranging from 0.69 to 0.75. For criterion A3 (problem solving 1), reliability was lower overall. Human-only agreement was moderate (ICC = 0.77), and ICCs under 0-shot prompting ranged from 0.68 to 0.71. Under 1-shot prompting, ICCs for A3 ranged from 0.68 to 0.76, indicating similar levels of agreement across prompting conditions.

More interpretive criteria showed different patterns. For criterion A4 (problem solving 2), reliability was excellent among human raters (ICC = 0.94) and remained high under 0-shot prompting, with ICCs between 0.89 and 0.92. Under 1-shot prompting, agreement for criterion A4 decreased, with ICCs ranging from 0.78 to 0.82. Criterion A5 (problem solving 3) showed perfect agreement among human raters (ICC = 1.00). When GPT-4o was included, ICCs remained high across all conditions, ranging from 0.79 to 0.94.

In contrast, criterion A6 (problem solving 4) exhibited the lowest levels of agreement. Human-only reliability was moderate (ICC = 0.60). When GPT-4o scores were included, ICCs under 0-shot prompting ranged from 0.43 to 0.46, and ICCs under 1-shot prompting ranged from 0.33 to 0.49. Across prompting conditions, agreement for A6 remained lower than for other criteria.

Across all criteria, ICC values varied only modestly across temperature settings, indicating that temperature had a limited effect on criterion-level reliability. In contrast, prompting approach (0-shot vs. 1-shot) and the nature of the scoring criterion exerted substantially stronger influences on agreement. Overall, agreement was higher for criteria involving well-defined conceptual knowledge and representational transformations, and lower for criteria requiring extended procedural execution and interpretive judgment.

Table 7. Criterion-level ICCs in Round 2

| Criterion | Human Only | Human & AI (GPT-4o), 0-shot | | | | Human & AI (GPT-4o), 1-shot | | | |
|---|---|---|---|---|---|---|---|---|---|
| | | T = 0.3 | T = 0.5 | T = 0.7 | T = 1.0 | T = 0.3 | T = 0.5 | T = 0.7 | T = 1.0 |
| A1 | 0.93 | 0.91 | 0.92 | 0.91 | 0.92 | 0.83 | 0.82 | 0.82 | 0.80 |
| A2 | 0.89 | 0.85 | 0.82 | 0.83 | 0.83 | 0.75 | 0.74 | 0.69 | 0.72 |
| A3 | 0.77 | 0.70 | 0.69 | 0.68 | 0.71 | 0.68 | 0.73 | 0.73 | 0.76 |
| A4 | 0.94 | 0.91 | 0.92 | 0.89 | 0.91 | 0.82 | 0.82 | 0.79 | 0.78 |
| A5 | 1.00 | 0.88 | 0.94 | 0.84 | 0.81 | 0.79 | 0.79 | 0.82 | 0.81 |
| A6 | 0.60 | 0.45 | 0.43 | 0.44 | 0.46 | 0.33 | 0.40 | 0.35 | 0.49 |

*Note.* T = temperature. A1 = physics knowledge; A2 = math skill; A3 = problem solving 1; A4 = problem solving 2; A5 = problem solving 3, A6 = problem solving 4.

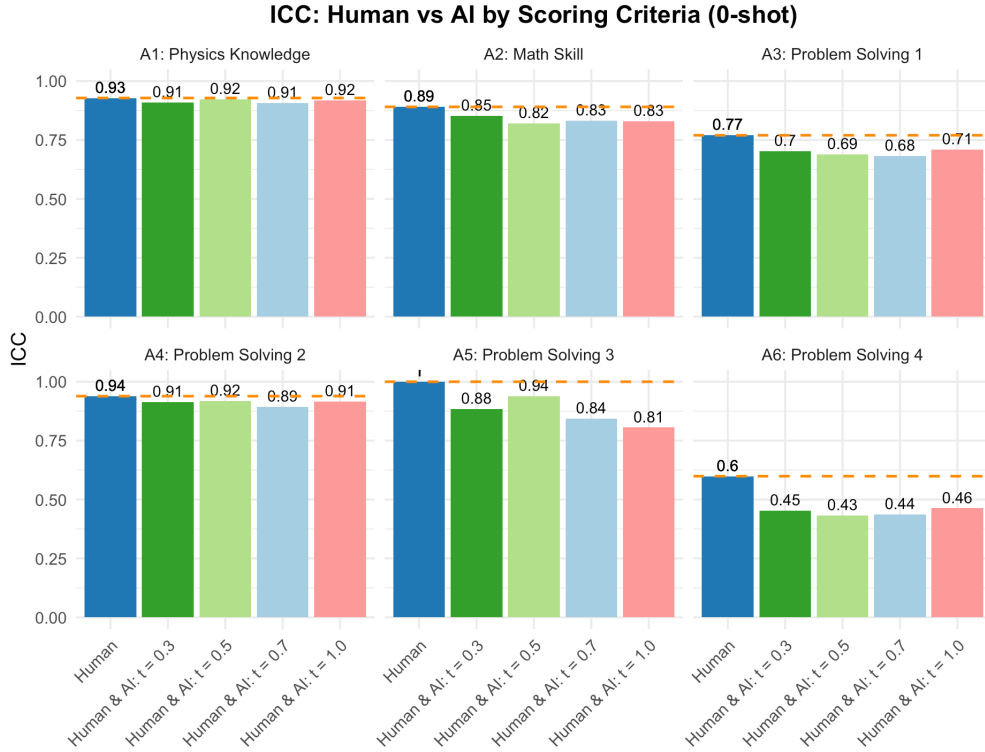

Figure 7. ICCs under 0-shot Prompting in Round 2

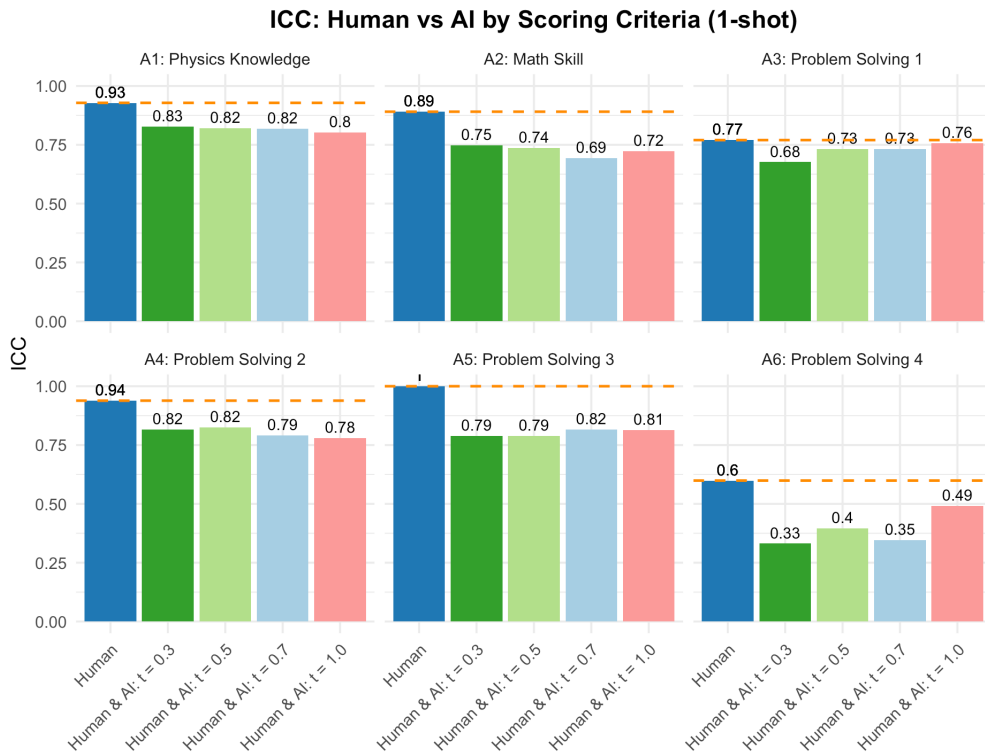

Figure 8. ICCs under 1-shot prompting in Round 2

## 5. Discussion

This study examined the reliability of generative AI–assisted scoring for physics constructed responses under different rubric structures, prompting formats, and temperature conditions. Across two rounds of scoring, results show that GPT-4o produces scores that align closely with human ratings when student responses are unambiguous (either clearly demonstrating mastery or clearly reflecting minimal understanding) and when scoring criteria are explicitly defined. In contrast, reliability was consistently lower for responses involving partial or ambiguous reasoning, both among human raters themselves and in comparisons between human raters and GPT-4o. These patterns indicate that generative AI largely reflects existing sources of scoring difficulty in STEM assessment, particularly those associated with partial-credit and interpretive judgments, rather than introducing fundamentally new forms of inconsistency.

### 5.1 Overall Human–AI Agreement at the Total and Criterion Levels

Results show that generative AI achieved levels of agreement with instructors that were comparable to human inter-rater reliability under several conditions. In Round 1, overall human–AI ICCs closely matched human-only ICCs, indicating that the model generally reproduced expert judgments when applying the holistic skill-based rubric. A similar pattern was observed in Round 2 under 0-shot prompting, where overall agreement remained in the good-to-excellent range.

At the criterion level, agreement was consistently higher for skills linked to well-defined conceptual knowledge and standard representational procedures, such as recognizing kinematic relationships or converting between vector forms. In contrast, lower agreement was observed for criteria requiring extended procedural execution or interpretive judgment. These results indicate that AI-assisted scoring is most reliable when rubric criteria are anchored to explicit, rule-based evidence. When criteria require holistic integration of multiple reasoning components, consistent evaluation becomes more difficult for both human raters and AI systems.

These findings show that generative AI can function as a stable extension of expert judgment for clearly specified scoring dimensions. Under these conditions, AI-assisted scoring may support preliminary evaluation and formative feedback without substantially altering established patterns of human assessment.

**5.2 Performance-Level Differences in Human–AI Agreement**

Across both rounds, reliability was consistently highest for responses at the extreme ends of the performance distribution. For clearly correct and clearly incorrect responses, both human raters and GPT-4o assigned similar scores, reflecting the presence of unambiguous scoring evidence. In contrast, agreement declined substantially for mid-level responses reflecting partial understanding or mixed reasoning.

In Round 1, both human-only and human–AI ICCs were low for this group, indicating substantial interpretive variation under holistic skill scoring. This variability reflects the inherent difficulty of assigning partial credit when multiple reasoning components are present but unevenly developed. In Round 2, the checklist-based rubric improved agreement for mid-level responses under 0-shot prompting. By providing explicit anchors for partial-credit decisions, the rubric reduced some sources of interpretive disagreement. However, this improvement was not consistent across all LLM settings. Under 1-shot prompting, agreement for mid-level responses deteriorated sharply, despite high human-only reliability. This pattern indicates that mid-level responses are particularly sensitive to how scoring criteria are operationalized and how models are guided during evaluation.

These findings demonstrate that partial-credit responses represent the primary source of instability in both human and AI scoring. Performance level should therefore be treated as a central consideration in the design and evaluation of AI-assisted assessment systems.

**5.3 Effects of Rubric Structure, Prompting Format, and Temperature**

Comparisons across rounds indicate that rubric structure had the strongest effect on scoring reliability. The transition from holistic skill scoring in Round 1 to checklist-based scoring in Round 2 improved human consistency and strengthened alignment with GPT-4o under 0-shot prompting. By linking scores to observable features of student work, the checklist rubric reduced ambiguity in partial-credit decisions and provided clearer evaluative anchors.

Prompting format also had a substantial impact. Under 0-shot prompting, AI-generated scores were stable and closely aligned with human ratings across temperature settings. Under 1-shot prompting, agreement declined, particularly for mid-level responses. Because the exemplar represented a fully correct solution, it likely biased the model toward complete correctness and reduced sensitivity to partial-credit distinctions.

Temperature settings had comparatively modest effects. Although higher temperatures increased variability in individual AI outputs for some responses, overall ICC values remained relatively stable. This suggests that sampling stochasticity plays a smaller role in scoring reliability than rubric specification and prompt design.

Overall, these results indicate that rubric structure functions as the primary determinant of AI scoring reliability, with prompting format acting as a secondary influence and temperature serving mainly as a fine-tuning parameter.

**5.4 Skill Representation and CDM-Inspired Rubric Design**

The observed patterns can be interpreted through the lens of CDM, which conceptualizes problem solving as the coordinated application of multiple latent skills. In CDM-based approaches, performance is interpreted with reference to an explicit skill structure that links observable response features to underlying competencies. When this structure is made explicit in scoring rubrics, it provides a shared interpretive framework for both human raters and automated systems.

In the present study, the more fine-grained checklist-style rubric more closely approximated this CDM logic by mapping individual skills to concrete, part-specific indicators. This mapping reduced the need for raters and the AI model to infer latent competence from loosely defined evidence. Instead, judgments were grounded in observable features of student work, leading to more stable scoring.

By contrast, holistic skill scoring required evaluators to integrate dispersed and sometimes inconsistent evidence across the full response. This process relies heavily on tacit professional judgment and contextual interpretation. Because LLMs lack domain-specific representations of physics problem solving, they must approximate such judgments using surface cues and statistical regularities. This limitation helps explain why agreement declined for skills involving extended reasoning and procedural monitoring.

From this perspective, rubric design serves as representational infrastructure for AI-assisted assessment. Well-specified skill mappings reduce interpretive burden, whereas underspecified mappings shift that burden onto raters and models. CDM-inspired rubrics therefore function not only as scoring tools but also as interfaces that shape how student reasoning is evaluated.

**5.5 Implications for AI-Assisted Assessment Design**

The findings yield several practical implications for the design and implementation of AI-assisted scoring in STEM education. First, rubric design should be prioritized. Analytic rubrics with explicit, observable criteria are essential for reliable AI-assisted scoring. Before tuning prompting strategies or model parameters, instructors and assessment designers should focus on developing structured rubrics that minimize interpretive ambiguity. Second, AI-assisted scoring is most appropriate for responses with clear evidence of mastery or non-mastery. For such responses, AI can support automated triage, preliminary scoring, and rapid feedback. Responses involving partial understanding should be treated as high-risk cases requiring human review.

Third, exemplar-based prompting should be used cautiously. When exemplars represent only fully correct solutions, they can distort model calibration and reduce sensitivity to partial credit. If exemplars are used, they should span a range of performance levels. Fourth, AI scoring should be embedded in hybrid workflows that preserve human oversight. Disagreement between human and AI scores should be interpreted as a diagnostic signal rather than as system failure, prompting closer inspection of ambiguous responses or unclear rubric criteria.

Finally, although this study focused on physics, the underlying challenges (e.g., partial-credit evaluation, multi-step reasoning, and heterogeneous response representations) are common across STEM disciplines. The design principles identified here are therefore likely to generalize to related domains such as engineering, mathematics, and chemistry.

### 5.6 Limitations and Future Directions

Several limitations should be considered in relation to the scope and aims of this study. The analysis focused on a carefully selected set of authentic responses from a single instructional context in order to support fine-grained investigation of rubric design and LLM configurations, including prompting format and temperature settings. While this design strengthens internal validity, broader generalization will require replication across additional courses and disciplines. In addition, the study examined one multimodal language model to isolate the effects of rubric structure and prompting strategies. Future work should evaluate whether similar patterns hold across alternative model architectures and deployment settings.

## 6. Conclusion

This study examined the reliability of generative AI–assisted scoring of undergraduate physics constructed responses under different rubric structures and LLM configurations. GPT-4o showed agreement with experienced instructors comparable to human inter-rater reliability for clear, extreme-performing responses scored with explicitly specified criteria, but agreement declined for mid-level responses involving partial or mixed reasoning.

Across conditions, rubric design emerged as the primary determinant of scoring reliability. The more fine-grained checklist-based skill rubric with explicit, observable criteria improved consistency for both human raters and the AI model, whereas holistic or loosely specified criteria increased interpretive variability. Prompting format exerted a secondary influence, and temperature settings had relatively limited effects. These findings indicate that effective AI-assisted assessment depends primarily on careful rubric design and well-defined hybrid human–AI workflows, rather than on model parameter tuning alone.

*Science Teaching: The Official Journal of the National Association for Research in Science Teaching*, *46*(1), 27-49. https://doi.org/10.1002/tea.20265

Alikaniotis, D., Yannakoudakis, H., & Rei, M. (2016). *Automatic text scoring using neural networks*. In *Proceedings of the 54th Annual Meeting of the Association for Computational Linguistics (Volume 1: Long Papers)* (pp. 715–725). Association for Computational Linguistics. https://doi.org/10.18653/v1/P16-1068

Attali, Y., & Burstein, J. (2006). Automated essay scoring with e-rater® V.2. *The Journal of Technology, Learning and Assessment*, 4(3). Retrieved from https://ejournals.bc.edu/index.php/jtla/article/view/1650

Brookhart, S. M. (2013). *How to create and use rubrics for formative assessment and grading*. Ascd.

Bui, N. M., & Barrot, J. S. (2025). ChatGPT as an automated essay scoring tool in the writing classrooms: how it compares with human scoring. *Education and Information Technologies*, *30*(2), 2041-2058. https://doi.org/10.1007/s10639-024-12891-w

Caraeni, A., Scarlatos, A., & Lan, A. (2024). Evaluating GPT-4 at Grading Handwritten Solutions in Math Exams. *arXiv preprint arXiv:2411.05231*.

Deane, P. (2013). On the relation between automated essay scoring and modern views of the writing construct. *Assessing Writing*, *18*(1), 7-24. http://dx.doi.org/10.1016/j.asw.2012.10.002

Floridi, L., & Chiriatti, M. (2020). GPT-3: Its nature, scope, limits, and consequences. *Minds and machines*, *30*(4), 681-694. https://doi.org/10.1007/s11023-020-09548-1

Holmes, N. G., & Wieman, C. E. (2016). Examining and contrasting the cognitive activities engaged in undergraduate research experiences and lab courses. *Physical Review Physics*
34

Turpin, M., Michael, J., Perez, E., & Bowman, S. (2023). Language models don't always say what they think: Unfaithful explanations in chain-of-thought prompting. *Advances in Neural Information Processing Systems*, *36*, 74952-74965.

Wang, Z., & von Davier, A. A. (2014). Monitoring of scoring using the e-rater® automated scoring system and human raters on a writing test. ETS Research Report Series, 2014(1), 1-21. https://doi.org/10.1002/ets2.12005

Williamson, D. M., Mislevy, R. J., & Bejar, I. I. (Eds.). (2006). *Automated scoring of complex tasks in computer-based testing*. Mahwah: Lawrence Erlbaum Associates.

Wu, M., & Aji, A. F. (2025, January). Style over substance: Evaluation biases for large language models. In *Proceedings of the 31st International Conference on Computational Linguistics* (pp. 297-312). https://aclanthology.org/2025.coling-main.21/

Yavuz, F., Çelik, Ö., & Yavaş Çelik, G. (2025). Utilizing large language models for EFL essay grading: An examination of reliability and validity in rubric-based assessments. *British Journal of Educational Technology*, *56*(1), 150-166.

Zhuo, J., Zhang, S., Fang, X., Duan, H., Lin, D., & Chen, K. (2024). ProSA: Assessing and understanding the prompt sensitivity of LLMs. *arXiv preprint arXiv:2410.12405*.
39

# Appendix
## Table A1. Rubric Version 1

| Rubric Category | Problem Parts | Mastery 3 | Growing 2 | Attempted 1 | No Attempt 0 |
|---|---|---|---|---|---|
| **Physics Knowledge** | | | | | |
| **A1:** Recognize that ignoring air resistance, objects moving under the influence of gravity can be analyzed with the approach of motion under constant acceleration | a, b, c | Consistently uses gravitational acceleration downward and no acceleration in horizontal direction | Sometimes recognizes acceleration as described in "Mastery" but sometimes makes other choice or fails to take advantage of this knowledge in some of the parts. | Attempted to use acceleration but didn't identifying downward component as g or horizontal as zero. | Did not make any use of acceleration concepts in the problem. |
| **A2:** Analyze the trajectory of a projectile (or a body undergoing motion at constant acceleration) based on its equations of motion. | a, b, c | Selected appropriate and relevant equations for all parts of the problem. | For some parts of the problem selected appropriate and relevant equations of motion, but not for all. | Selected equations of motion that were not appropriate or relevant for any part. | Did not attempt to make use of any constant acceleration equations of motion. |
| **Math Skills** | | | | | |
| **A3:** Convert between magnitude-angle and component descriptions of a vector. | a, b | Converted from magnitude and angle to components and back again as needed. | Correctly converted sometime, but not in every case. | Recognized that the problem involved concepts of magnitude-angle and components, but didn't apply the relationships correctly. | Didn't use the relationship between vector magnitude-angle and components in any part. |
| **Problem Solving** | | | | | |
| **A4:** Extract useful information from problem description | all | All of the relevant information has been extracted from the problem statement | One or two pieces of information were not extracted or used, making it difficult to solve the problem. | Most of the relevant information from the problem statement was not recognized. | The problem was not attempted. |
| **A5:** Clearly identify which physics principles or equations are being applied to solve the problem. | all | When solving, the solution always started with a clear statement of which physics principle and/or equation from the equation sheet is the starting point. | For some equations, it was not clear which physics principle and/or equation from the equation sheet served as the starting point. | None of the work was explicitly connected to physics principles or equations on the equation sheets. | The problem was not attempted. |

| | | A6: Mathematical Procedures | all | Consistently, the work is written out clearly and no math errors occur. | The work is mostly written out clearly but there are one or more math errors. | The work is mostly or entirely absent. There is no way to understand how the final answer is connected to the initial information in the problem. | The problem was not attempted. |

Table A2. Rubric Version 2

| Skill | Detailed Criteria | Part (a) checklist | Max points | Part (b) checklist | Max points | Part (c) checklist | Max points |
|---|---|---|---|---|---|---|---|
| **A1: Physics Knowledge** | Use appropriate kinematic equations to analyze the motion of objects in one, two, or three dimensions, under constant or non-constant acceleration. | Used eq. of motion for y in vert. direction with constant a | 1 | Used v eq. of motion with constant a in vert. direction and constant v in the horiz. direction | 2 | used x eq. Of motion with constant v in the horiz. Correctly interpreted whether object landed in pool or not based on the calculated x | 1 |
| **A2: Math Skills** | Convert between magnitude-angle and component descriptions of a vector. | Found vy=vsin(theta) | 1 | Found vy=vsin(theta) and vx=vcos(theta), found mag used tan correctly to find direction | 3 | vx=vcos(theta) | 0 |
| **A3: Problem Solving 1** | Extract useful information from problem description. | v=20, theta=30, y0=40m, find t at y=0 | 4 | v=20, theta=30 | 0 | v=20, theta=30, location of the pool | 1 |
| **A4: Problem Solving 2** | Identify which physical principles apply to a given problem. | Identified that a=g (free fall along y) | 1 | Recognize that a=g in vert. direction and a=0 in horiz. direction | 2 | identified that a=0, i.e. vx constant | 1 |
| **A5: Problem Solving 3** | Convert among multiple representations of a situation (verbal, algebraic, numerical, graphical, pictorial). | N/A - No criteria for this section | N/A | N/A - No criteria for this section | N/A | correctly interprets answer (insider or outside pool depending on x(t)) | 1 |
| **A6: Problem Solving 4** | Prior math skills. Ability to correctly carry out algebra/calculation | correctly solves for root of a 2nd deg polynomial | 1 | numerical calculations correct? | 1 | numerical calculations correct correctly plugs t into x(t) | 0 |